\documentclass[10pt,twocolumn,letterpaper]{article}

\usepackage{cvpr}
\usepackage{times}
\usepackage{epsfig}
\usepackage{graphicx}
\usepackage{amsmath}
\usepackage{amssymb}

\usepackage{algorithm}
\usepackage{algorithmic}
\usepackage{multirow}
\usepackage{amsmath}
\usepackage{xcolor}
\usepackage{colortbl}
\definecolor{mygray}{gray}{.9}

\usepackage[pagebackref=true,breaklinks=true,letterpaper=true,colorlinks,bookmarks=false]{hyperref}

 \cvprfinalcopy 

\def\cvprPaperID{1682} 

\ifcvprfinal\fi
\begin{document}

\title{SOT for MOT}

\author{Qizheng He\thanks{Work was done during an internship at Megvii Research.}\\
IIIS, Tsinghua University\\
Beijing, China\\
{\tt\small hqz14@mails.tsinghua.edu.cn}
\and
Jianan Wu\\
Megvii Inc. (Face++)\\
Beijing, China\\
{\tt\small wjn@megvii.com}
\and
Gang Yu\\
Megvii Inc. (Face++)\\
Beijing, China\\
{\tt\small yugang@megvii.com}
\and
Chi Zhang\\
Megvii Inc. (Face++)\\
Beijing, China\\
{\tt\small zhangchi@megvii.com}
}
\maketitle

\begin{abstract}
   In this paper we present a robust tracker to solve the multiple object tracking (MOT) problem, under the framework of tracking-by-detection. As the first contribution, we innovatively combine single object tracking (SOT) algorithms with multiple object tracking algorithms, and our results show that SOT is a general way to strongly reduce the number of false negatives, regardless of the quality of detection. Another contribution is that we show with a deep learning based appearance model, it is easy to associate detections of the same object efficiently and also with high accuracy. This appearance model plays an important role in our MOT algorithm to correctly associate detections into long trajectories, and also in our SOT algorithm to discover new detections mistakenly missed by the detector. The deep neural network based model ensures the robustness of our tracking algorithm, which can perform data association in a wide variety of scenes. We ran comprehensive experiments on a large-scale and challenging dataset, the MOT16 benchmark\cite{milan2016mot16}, and results showed that our tracker achieved state-of-the-art performance based on both public and private detections.
\end{abstract}


\section{Introduction}
Multiple object tracking is the problem of automatically identifying multiple objects in a video and representing them as a set of trajectories with high accuracy. It is an important problem in computer vision because it can play a fundamental role in various applications, \eg in automatic driving and surveillance video processing. In this paper we mainly focus on pedestrian tracking in video, but with a general object detector, our method can be easily generalized to deal with general types of objects.

Various ways have been proposed to solve the multiple object tracking problem. However, thanks to the rapid development of deep learning based object detection methods, most of recent state-of-the-art researches has focused on the framework of tracking-by-detection. After getting all detection hypotheses in the video, the tracking problem then becomes a data association problem to combine detections of the same object into a corresponding trajectory.

Though during the past years more and more advanced tracking-by-detection algorithms have been proposed, we observe there are still some parts in the framework that need improvement. First, we notice that the tracking results heavily rely on the quality of detections. However, detectors may fail to detect objects in some moments, \eg when objects are in crowded scenes or when they are partially occluded. This problem cannot be completely solved by nowaday detectors, and is difficult all the way because the detector treats each frame of the video separately. Few previous tracking-by-detection trackers try to solve this problem, so when the detector fails, most tracking algorithms are only possible to find those missing detections by coarse methods like interpolation within trajectories, which needs improvement. Second, a reliable appearance model is crucial to data association. Most previous works provide hand-crafted affinity measurements as their appearance models, which may not be robust enough.

Based on the tracking-by-detection framework, this paper makes the following contributions:  First, in order to further reduce the number of false negatives generated by the detector, we innovatively use single object tracking within the framework of multiple object tracking, which to our knowledge is the first to explore this method. Second, instead of a hand-crafted appearance model used in many previous works, we use a deep-neural network based appearance model throughout the whole framework, which can provide robustness in complex tracking scenes. Our experiments show with the help of our appearance model, we can pay little attention to spatial-temporal constraints on tracking problems, and still achieve state-of-the-art performance by a reliable affinity measure based on appearance.

\begin{figure*}
\begin{center}
\includegraphics[width=.8\linewidth]{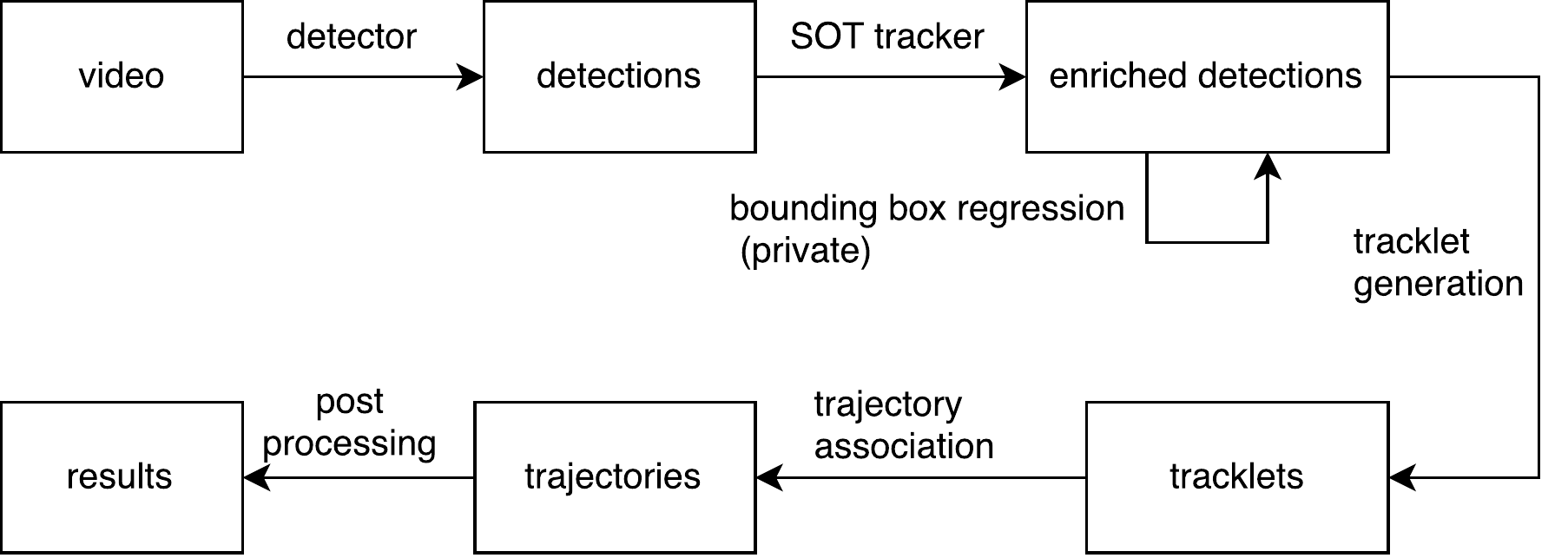}
\end{center}
\caption{An overview of our tracking framework.}
\label{fig:framework}
\end{figure*}

\section{Related work}
Various solutions have been proposed for the MOT problem. One possible way is the recurrent neural network (RNN). Milan \etal \cite{milan2016online} provided an end-to-end deep learning approach using RNN, which can predict target existence states and perform data association in a whole network. However, though attracting as it sounds, their results are currently still not comparable with other state-of-the-art algorithms in performance. There are also other approaches, like MCMC-based particle filtering \cite{choi2013general,khan2005mcmc}.

Recent researches have focused on the tracking-by-detection framework, which has the following advantages:
a) it can automatically detect new objects when they appear in the scene, and when they leave the scene their detections naturally disappear,
b) because of the detector treats each frame in the video separately, camera motion will have little impact on detection hypotheses generation in the first place, and
c) with strong category information given by the detector, it can effectively prevent bounding boxes from drifting to the background.

An example of such approach is bipartite matching. It uses nodes in a graph to represent detections, uses edge weights to represent affinity between two detections, and then find the optimal matching between two consecutive frames. Its advantages are obvious: it is efficient because the Kuhn-Munkres algorithm proposed by Kuhn and Harold \cite{kuhn1955hungarian} for bipartite matching runs in polynomial time, and it's an online algorithm. However, online tracking algorithms have their limits, and as Choi \cite{choi2015near} pointed out, global or batch tracking methods have advantage over online methods by optimizing through a larger number of consecutive frames.

As a more advanced version, Zhang \etal \cite{zhang2008global} viewed the MOT problem as an MAP data association problem, and optimized using min-cost network flow in a global way. The algorithm for min-cost network flow is also polynomial. Though under their model the network flow solution is optimal, the model itself has limitations. For example, it only considers affinity between consecutive detections of one object and does not take higher order relationship within a trajectory into account. Another weakness is though they proposed an occlusion model, their method can hardly recover the trajectory if the object suffers a relatively long time of occlusion. After their work, Pirsiavash \etal \cite{pirsiavash2011globally} proposed an algorithm that can find a sub-optimal solution even faster using dynamic programming, and Butt and Collins \cite{butt2013multi} considered trajectory smoothness constraints by introducing lagrangian relaxation to min-cost network flow.

There are more complex models based on the framework of tracking-by-detection. Milan \etal \cite{milan2014continuous} made a more complete problem formulation for tracking and defined a complicated objective function. Though hard to find the optimal solution, they showed a sub-optimal solution can outperform optimal solutions in weaker models. \cite{yang2011learning,yang2012online} also view the MOT problem in an energy minimizing aspect, and solved using CRF models. \cite{hong2016online,wen2014multiple} provided other ways to design data association methods.

Some other researchers consider all pairwise affinities between detections in a trajectory, representing each trajectory as a clique. The MOT problem then becomes a multi-clique problem or equivalently a graph partitioning problem or multicut problem, similar detections of the same object can be clustered spatially and temporally. \cite{dehghan2015gmmcp,keuper2016multi,ristani2014tracking,tang2015subgraph,tang2016multi,zamir2012gmcp} are examples of this approach. The multi-clique problem can be solved optimally by binary integer programming \cite{ristani2014tracking}, or find a nearly optimal solution by heuristic approaches \cite{tang2015subgraph}. There are other methods that introduced high order terms in a graph model, like \cite{choi2015near}, which used a CRF model to find the solution for their graphical model. These methods also achieved high performance.

The problem of single object tracking is different from multiple object tracking. A common problem formulation of SOT is to find a short-term trajectory of a single object (foreground), only the bounding box of the tracked object in the first frame is given, and the solution is always without re-detection. Other objects appear in the scene (even of the same category) are all considered as background. \cite{bertinetto2016staple,liu2016structural,nam2015learning,qi2016hedged,tao2016siamese,wang2015visual,wang2016stct,zhu2016beyond} and \etc. proposed solutions for the SOT problem.

\section{Proposed framework}
\label{sec:Proposed framework}
Our whole tracking framework is illustrated in Fig. \ref{fig:framework}. Our algorithm runs like this: first we get all detections from the detector (Sec. \ref{sec:Detection}), then perform single object tracking on all detections (Sec. \ref{sec:SOT algorithm}), which results in an enlarged detection set $\{D'_1,D'_2,\cdots,D'_N\}$, next we perform our multiple object tracking algorithm on the new detection set (Sec. \ref{sec:MOT algorithm}), finally we perform a post-processing process (Sec. \ref{sec:Post processing}) and get the final tracking result.

\subsection{Notations}
\label{sec:Notations}
A video sequence of $N$ frames is notated as $V^N=\{F_1,F_2,\cdots,F_N\}$.
In each frame $F_i$ there is a set of detection hypotheses generated by the detector, represented as $D_i^n=\{d_1,d_2,\cdots,d_n\}$. A detection $d_i$ consists of its frame number $d_i[I]$, its score $d_i[s]$ and its bounding box $d_i[B]=(d_i[x],d_i[y],d_i[w],d_i[h])$, where $(d_i[x],d_i[y])$ is the top-left point of the bounding box, $d_i[w]$ is the width and $d_i[h]$ is the height. A trajectory is defined as a set of detections of the same object, which is notated as $t^L=\{d_1,d_2,\cdots,d_L\}$. Note that we do not have the constraint that $\{d_1,d_2,\cdots,d_L\}$ are from consecutive frames, for example a trajectory $t$ can represent an object's presence in frame $2,3$ and $5$, without its presence in frame $4$. A tracklet is informally defined as a short-term and continuous trajectory.

Let $I(d)$ be the image within detection $d$'s bounding box, our appearance model can calculate its feature $\hat{f}(I(d))$. To increase the robustness, let $I(\bar{d})$ be the horizontally flipped image within $d$'s bounding box, the feature of detection $d$ is calculated as
\begin{equation}
f(d)=\frac{1}{2}\left(\hat{f}(I(d))+\hat{f}(I(\bar{d}))\right).
\end{equation}
Two detections $d_1$ and $d_2$ have affinity score
\begin{equation}
\mathcal{D}(d_1,d_2)=||f(d_1)-f(d_2)||_2.
\end{equation}
$\mathcal{D}(d_1,d_2)$ is bounded in the range $[0,1]$.\\
A trajectory $t^L$'s feature is defined as the average feature of all its detections, i.e.
\begin{equation}
f(t^L)=\frac{1}{L}\sum_{d\in t^L}f(d).
\end{equation}
Trajectory $t$ and detection $d$ have affinity score
\begin{equation}
\mathcal{D}(t,d)=||f(t)-f(d)||_2.
\end{equation}


\subsection{Detection}
\label{sec:Detection}
Detection is the first step of the tracking-by-detection framework. To emphasize the importance of detection for tracking, we first introduce MOTA, a widely used measurement for multiple object trackers, whose definition in \cite{milan2016mot16} is as the following:
\begin{equation}
\text{MOTA}=1-\frac{\sum_t(\text{FN}_t+\text{FP}_t+\text{IDSW}_t)}{\sum_t \text{GT}_t}.
\end{equation}
Yu \etal \cite{yu2016poi} noticed that the sum of false positives (FP) and false negatives (FN) strongly affects the value of MOTA, which means the performance of a tracking algorithm is strongly based on the quality of detections.

\begin{figure}
\begin{center}
\includegraphics[width=\linewidth]{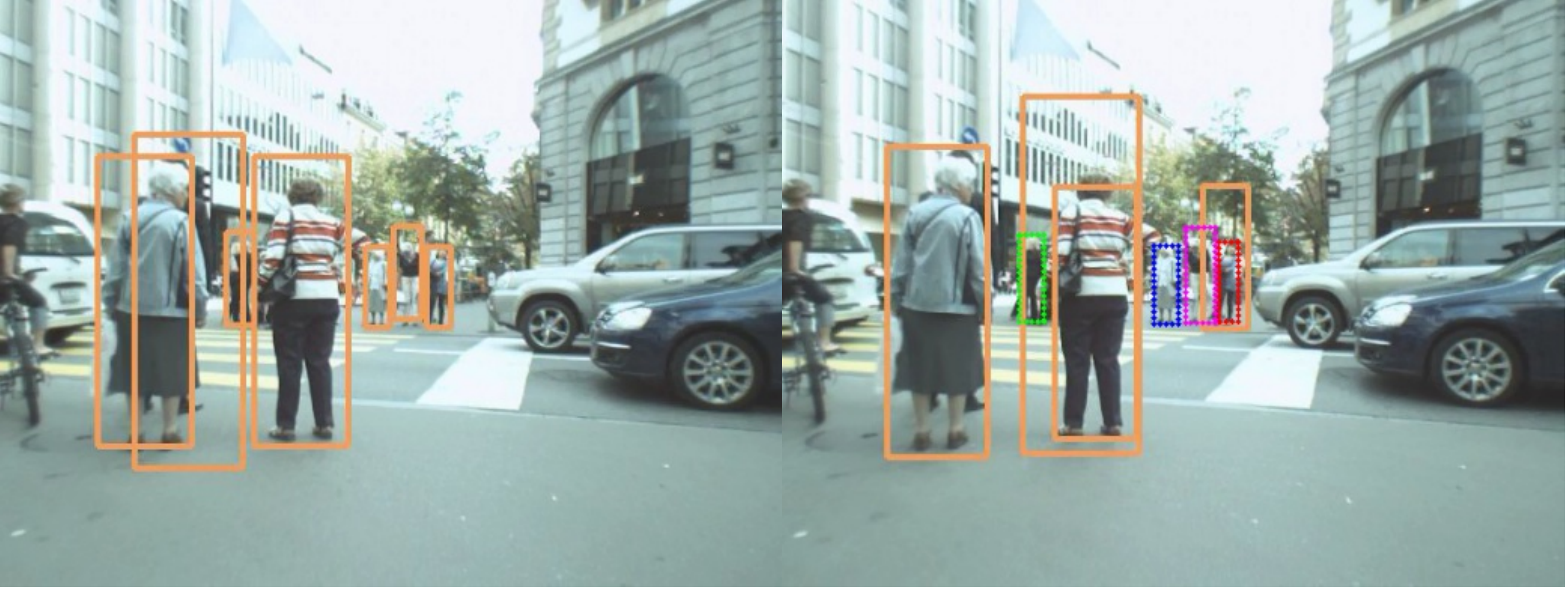}
\end{center}
   \caption{Public detections (bounding boxes in solid lines) in MOT16-05 \#15 (left), \#16 (right). By tracking detections in frame \#15, our SOT algorithm can help to find objects missed by the detector in frame \#16 (dotted bounding boxes), thus reducing the number of FNs. The figure is best shown in color.}
\label{Figure 2}
\end{figure}

A state-of-the-art detector with low FPs and FNs is of great help to the whole MOT algorithm, which is indicated by the definition of MOTA. However, detectors have their limitations. If we try to analyze the source of false negatives produced by detectors, when objects are crowded or partially occluded, they may fail to provide detections of objects, thus result in FNs. It's the tracker's duty to further reduce the number of FPs and FNs by a higher order of image and spatial-temporal information. In the next part, we present our SOT algorithm that can help to find some FNs in the detection. Fig \ref{Figure 2} gives an illustration for our SOT algorithm's application.

\subsection{SOT algorithm}
\label{sec:SOT algorithm}
There are plenty of works related to the study of single object tracking algorithms, however, to our knowledge there is no previous work gained success by directly using SOT algorithms to improve the performance of MOT algorithms.

One might think we can simply use SOT algorithms on each object in the MOT problem to get a solution for MOT. One reason that it cannot be easily used is model drift. It happens commonly when the recorded appearance of an object in the model slowly drifts away from its real appearance, and will finally cause the bounding box drift to the background, making it hard to terminate the trajectory. Another reason is it does not consider the interaction between objects, which is important when we have multiple objects to track. One more reason is sometimes we even do not know when to start tracking when the object appears for the first time.

Tao \etal \cite{tao2016siamese} present a single object tracker based on a siamese deep neural network and show with the robust matching function learned by the network, a simple tracker can achieve state-of-the-art on SOT benchmarks. Inspired by their work, we propose a variant of SOT algorithm as one step in our whole MOT framework, immediate after the detection step. The aim of this step is to further reduce the number of FNs in the detection, without significantly increase the number of FPs.\\

\noindent \textbf{Detection candidate sampling}
Here we design a function $next(B,d_{b},I)$. Suppose the detector gives a detection $d_{b}$ of an object $o$ (this implicitly assumes $d_{b}$ is a true positive, which is our assumption), we want to find a detection $d_{r}$ as the result, which corresponds to object $o$ on the frame with frame number $I$ (if $o$ really appears in that frame). $B$ is the provided prior knowledge of the location of $d_{r}$'s bounding box. We first generate a set $D$ containing detection candidates in the $I$-th frame, using dense sampling on the distribution of the location and scale of $d_{r}$'s bounding box:
\begin{equation}
D_{c}=\{d_{c}|d_{c}[I]=I,d_{c}[B]\sim \mathcal{N}(B,\sigma)\}.
\end{equation}

Note here our assumptions may be violated, \eg the time when we call the $next$ function with $d_{b}$ being a FP. The problems followed will be dealt with in our MOT algorithm.\\

\noindent \textbf{Detection candidate matching}
Using our deep appearance model introduced in Sec. \ref{sec:Appearance model}, we can find the best matched detection among all detection candidates $d_{c}\in D_c$, using the appearance similarity between $d_{c}$ and $d_{b}$:
\begin{equation}
d_{r}=\mathop{\arg\max}_{d_{c}\in D_{c}} \mathcal{D}(d_{c},d_{b}).
\end{equation}
Finally if $\mathcal{D}(d_{r},d_{b})$ is larger than a pre-defined threshold $\tau=0.5$, the function $next(B,d_{b},I)$ returns $d_{r}$ as the result. Otherwise it returns $\emptyset$, meaning that object $o$ does not appear in the $I$-th frame.

With a carefully trained appearance model, the detection candidate sampled with a more precise location will have appearance similarity score $\mathcal{D}$ larger than other candidates that deviated from the true location of the object. This property enables our SOT algorithm to find object in new frames with good localization accuracy.\\

\noindent \textbf{SOT with multiple object interaction}
Here we present the proposed SOT algorithm in algorithm \ref{algo}, where we have considered multiple object interactions, and it runs in two directions on time axis, within a batch of frames that contain detection set $D=\bigcup_i D_i$. We use $\tau_I=0.5$, $\tau_a=0.75$, $e=0.9$ and $L=15$.

\begin{algorithm}[htb]
\caption{proposed SOT algorithm}
\label{algo}
\begin{algorithmic}[1]
\STATE initialize $ends=\{(d,dir)|d\in D,dir\in \{-1,1\}\}$
\STATE initialize empty priority queue $Q$
\STATE $\forall d\in D$, put the backward tracklet end $(d,d,d[s],1,-1)$ and the forward tracklet end $(d,d,d[s],1,1)$ into Q
\WHILE{$Q$ not empty}
    \STATE extract a valid tracklet end $(d,d_{b},s,l,dir)$ with largest score $s$ from $Q$
    \IF{$dir=1$}
        \STATE \textbf{if} $l>=L$ \textbf{then} \textbf{continue}
        \IF {$\exists~(d_0,-dir)\in ends$ where $d_0[I]=d[I]+dir$ and $\mathcal{D}(d_0,d)>\tau_a$ and $IOU(d_0,d)>\tau_I$}
            \STATE merge two tracklet ends $d$ and $d_0$, update $ends$
            \STATE \textbf{continue}
        \ENDIF
        \STATE $d'=next(d[B],d_{b},d[I]+dir)$
        \IF{$\mathcal{D}(d_b,d')<\tau_a$ or $\exists d_0\in D$ where $IOU(d_0,d')>\tau_I$}
            \STATE \textbf{continue}
        \ENDIF
        \STATE $D=D\bigcup \{d'\}$
        \STATE merge $d'$ with current tracklet end
        \STATE insert $(d',d_{b},s\cdot e,l+1,dir)$ into Q
    \ELSE
        \STATE a similar process with direction being backward
    \ENDIF
\ENDWHILE
\end{algorithmic}
\end{algorithm}
After running this algorithm, we will get a set of tracklets within the batch, which may contain new detections found by the SOT algorithm.

Here are explanations of this algorithm. First, extending tracklet endpoints with decreasing order in score can naturally discover high-quality tracklets in the beginning, therefore effectively prevent FPs from forming longer tracklets. Second, the reason that we use a previously confident detection $d_{b}$ rather than $d$ as the standard appearance for the object is to prevent model drift, which commonly happens in traditional SOT algorithms. If $\mathcal{D}(d_b,d')<\tau_a$, we think the object is no longer visible.

This algorithm will certainly find new FPs that are similar to existing FPs in appearance, however it is not hard to discriminate them in the following MOT algorithm part.\\

\noindent \textbf{Bounding box regression}
We use bounding box regression on SOT results from private detection, and observe by doing this the number of FP generated by the SOT algorithm is significantly cut down. In tracking results from public detection we do not perform this step though it would bring significant improvement on MOTA, because we think doing this will make it unfair to compare with other tracking algorithms.\\


\noindent \textbf{Naive template matching implementation}
As an argument to show the ability of our SOT step, we tried a naive implementation using template matching instead of our deep neural network based appearance model. We used the function \emph{gpu::matchTemplate()} as an alternative appearance model, which is provided by opencv using the correlation between two patches of images. This affinity measure is similar to \cite{briechle2001template}, and is simple and fast. Our results show even with a low-quality matching function, performing the SOT step can still improve the performance of the whole MOT algorithm.

\subsection{MOT algorithm}
\label{sec:MOT algorithm}
In this part we introduce our proposed MOT framework. Our algorithm works in a batch, i.e. let $\tau$ be the batch size, we perform data association in a temporal window $[t_0,t_0+\tau)$. The batch size we use in our algorithm is related to the FPS of the video, \ie each batch has time length $1$ second. As the start, we first perform NMS on all detections generated by the SOT step.\\

\noindent \textbf{Tracklet generation}
Many of recent works used the pipeline of first generate consecutive tracklets then merge them into longer trajectories in their tracking framework. For example, Wang \etal \cite{wang2014tracklet} generate tracklets based on posteriori probabilities, and find the solution by successive shortest path algorithm. Choi \cite{choi2015near} generate tracklets by greedily find the detection that matches best with any detection within the current tracklet according to their appearance model, the ALFD metric, then merge that detection into the tracket. \cite{huang2008robust,prokaj2011inferring,wang2016tracklet} are other works that introduced tracklet generation. An advantage of this pipeline is it naturally solves the occlusion problem, with the help of a powerful appearance model that can link spatial-temporal separated tracklets of the same object together.

Now we present our tracklet generation method. Let $T_{id}^L=\{t_1,\cdots,t_L\}$ be the set of tracklets formed before time $id$. We construct a bipartite graph $G(V,E)$, each node $v\in V_0$ represent a tracklet $t\in T_{id}$, each node $v\in V_1$ represent a detection hypotheses $d\in D_{id}$, and $V=V_0\cup V_1$. The edge weight between $t$ and $d$ is defined as
\begin{equation}
dist(t,d)=\mathcal{D}(tail(t),d)
\end{equation}
where $tail(t)$ contains the last $5$ detections of $t$.

We perform bipartite matching on $G$ using Kuhn-Munkres algorithm \cite{kuhn1955hungarian} and get a new set of tracklets $T_{id+1}$ by merging the matched tracklets and detections. $T_{id+1}$ also contain those unmatched detections by regarding them as tracklets of length $1$. By letting $id=t_0,t_0+1,\cdots,t_0+\tau-1$ we repeatedly perform bipartite matching in the batch. Finally we delete tracklets with length $\leq l_{min}$ ($l_{min}=1$).\\

\noindent \textbf{Long-term trajectory association}
After the tracklet generation step, we will gets a set of trajectories $T$. We repeatedly find trajectory $t_1$ and $t_2$ with the maximum appearance affinity which satisfy the association constraints $A(t_1,t_2)$, until there are no such pairs:
\begin{equation}
(t_1,t_2)=\mathop{\arg\max}_{t_1'\neq t_2'\in T,A(t_1',t_2')=1}\mathcal{D}(t_1',t_2').
\end{equation}
We merge $t_1$ and $t_2$ into a new trajectory, if they temporally overlap on frame $I$ then we choose the detection with larger score on $I$. This step heavily relies on the powerful ability of our appearance model.\\

\noindent \textbf{Association constraints}
Now we introduce our association constraints $A(t_1,t_2)$ for two trajectories $t_1$ and $t_2$:\\
1) IOU constraint:
$t_1$ and $t_2$ may overlap on some time segments. Suppose in one time segment $[a,a+n]$ there are detections $d_0,\cdots,d_n\in t_1,~d_0',\cdots,d_n'\in t_2$ where $d_i[I]=d_i'[I]=a+i,~\forall~0\leq i\leq n$. The IOU constraint requires for each overlapping time segment the average IOU of detections should be larger than a threshold $\tau=0.5$, i.e.
\begin{equation}
\frac{1}{n+1}\sum_{i=0}^n IOU(d_i,d_i')>\tau.
\end{equation}
2) Suppose in the trajectory $t_1\cup t_2$ there's a gap $(a,a+n)$, \ie $\exists d_1,d_2\in t_1\cup t_2$ where $d_1[I]=a,d_2[I]=a+n,~\forall a<i<a+n,~\nexists d\in t_1\cup t_2~s.t.~d[I]=i$. We think in a video of a moving scene the velocity estimation is unreliable, therefore we use a weak spatial-temporal constraint: Let $v$ be the estimated velocity in the gap using linear velocity assumption, then $v$ should satisfy
\begin{equation}
v\leq
\begin{cases}
w & l=1\\
\frac{2}{3}w & 1<l\leq 5\\
\frac{1}{3}w & l>5\\
\end{cases}
\end{equation}
where $w=\frac{1}{2}(d_1[w],d_2[w])$.\\
3) Gap length constraint: all gaps within $t_1\cup t_2$ should have length less or equal than $l_{max}=60$.
%

\subsection{Post processing}
\label{sec:Post processing}


%
%
%
%

\noindent \textbf{Interpolation}
The trajectories generated by the previous steps may consist of several continuous parts separated by temporal gaps, which commonly happens when an object is occluded for some time or the detector missed detections. If occlusion occurs, our algorithm has the ability to link trajectories of that object before and after the gap based on our powerful appearance model. We use linear velocity assumption to interpolate detections within the gap.\\

\noindent \textbf{Smooth}
We also adopt refinements based on smoothness. Let $t$ be a trajectory, and detection $d\in t$, we use detections in $t$ whose frame number within a sliding temporal window $[d[I]-l,d[I]+l]$ to revise the location of $d$ (we set $l=3$). This step can make the trajectory more smooth and refine those outliers mislinked into the trajectory.

\section{Appearance model}
\label{sec:Appearance model}
A high-accuracy appearance model is fatal to designing a robust tracker. An ideal appearance model should have the ability of determining whether two detections in two frames within a video is the same object (providing pairwise affinity measure). If the image within two detections $d_1$ and $d_2$ are from the same object then the appearance model should give an affinity value $\mathcal{D}(d_1,d_2)$ close to $1$, otherwise it should give $\mathcal{D}(d_1,d_2)$ close to $0$. However, relatively less literatures focused on the design of appearance models.

In previous works most researchers used hand-crafted features as their appearance models. \cite{andriyenko2012discrete,godec2013hough,kuo2010multi} used HOG-features introduced by Dalal and Triggs \cite{dalal2005histograms}, and \cite{kuo2010multi} also used color histograms. Choi \cite{choi2015near} proposed the aggregated local flow descriptor (ALFD) as their appearance model, originated from optical flow and interest point trajectories. Relatively few literatures, like \cite{tang2016multi}, used deep learning methods to design matching functions. We propose a deep learning based appearance model which can lead to a better performance in more complex scenes. With sufficient training data, our model can automatically learn a generic function to produce a score of similarity using two detections' appearance, which is robust to partial occlusion, illumination change, angle of view difference, scale variation and other appearance distortions.

Now we introduce the implementation of our appearance model. The details of the model is similar to \cite{liu2016end}. To be concrete, the image patch for feature extraction will be first resized to $192\times 64$, then used as the input of our network. In the first stage, we use a ResNet \cite{he2015deep} structure as our dCNN component to train a classification task for objects. In the second stage, the first $34$ layers extracted by the dCNN component is then used to pass an LSTM based RNN component to extract the feature $\hat{f}(d)$ of object $d$, which has dimension $1024$. The appearance affinity score of two detections $d_1$ and $d_2$ is calculated as the L2 distance of the features, which is mentioned before in \ref{sec:Notations}.

For training, we use publicly available datasets, Market1501 \cite{zheng2015scalable} for the first classification stage and CUHK03 \cite{li2014deepreid} for the second stage. We use triplet loss \cite{schroff2015facenet} as our loss function in the second stage, which can ensure similar detections of the same object can have a high affinity score, and detections from different objects can have a low affinity score.


\section{Experiments}
\label{sec:Experiments}
We tested our tracking algorithm on the MOT16 benchmark\cite{milan2016mot16}. It is a collection of existing and new data (part of the sources are from \cite{leal2015motchallenge} and \cite{ess2007depth}), containing $14$ challenging real-world videos of both static scenes and moving scenes, $7$ for training and $7$ for testing. It is a large-scale dataset, composed of totally $110407$ bounding boxes in training set and $182326$ bounding boxes in test set.
All video sequences are annotated under strict standards, their ground-truths are highly accurate, making the evaluation meaningful. For the evaluation, the CLEAR MOT metrics \cite{bernardin2008evaluating} is used. Note that the MOT16 benchmark carefully annotated some ``ignore'' classes, detecting or not detecting them will not affect the result of evaluation.

\subsection{Detections used}
For public detection, we used DPM v5 provided by Felzenszwalb \etal \cite{felzenszwalb2010object} as our public detection. The MOT16 Challenge has already officially presented the public detection results with detection score larger than $-1$, making it fair to compare between public trackers. We perform NMS-IOM on the detections, where the intersection over minimum (IOM) of two detections $d_1$ and $d_2$ is defined as
\begin{equation}
\mathit{IOM}(d_1,d_2)=\frac{|d_1\bigcap d_2|}{\min(|d_1|,|d_2|)}
\end{equation}
in \cite{dollar2009integral}. As Tang \etal \cite{tang2015subgraph} pointed out, on public detections performing NMS with IOM is more useful than with IOU due to the property of DPM v5 \cite{felzenszwalb2010object}. We use threshold $0.6$ for NMS-IOM in practice.

For private detection, Yu \etal \cite{yu2016poi} introduced their high-performance detector based on Faster R-CNN, and made their private detection results publicly available. We used their private detections with detection score larger than $0.3$, except for the video MOT16-04, for which we use threshold $0.1$. The NMS-IOU threshold is $0.6$. The above setting is the same as \cite{yu2016poi}. Note the performance of the private detector is obviously significantly better than the public detector.

\cite{yu2016poi} provided a table for detection performance evaluation on both public and private detectors.

\subsection{Appearance model analysis}
\begin{table*}
\begin{center}
\begin{tabular}{|c|c|c|c|c|c|c|c|c|c|c|c|}
\hline
Method & MOTA$\uparrow$ & MOTP$\uparrow$ & FAF$\downarrow$ & MT$\uparrow$ & ML$\downarrow$ & FP$\downarrow$ & FN$\downarrow$ & ID Sw.$\downarrow$ & Frag$\downarrow$ & Hz$\uparrow$\\
\hline
NOMT \cite{choi2015near} & \textbf{46.4} & \textbf{76.6} & 1.6 & 18.3\% & 41.4\% & 9753 & \textbf{87565} & \textbf{359} & \textbf{504} & 2.6\\
\hline
JMC \cite{tang2015subgraph} & 46.3 & 75.7 & 1.1 & 15.5\% & \textbf{39.7\%} & 6373 & 90914 & 657 & 1114 & 0.8\\
\hline
\textbf{SOT+MOT} & 44.7 & 75.2 & 2.1 & \textbf{18.6\%} & 46.5\% & 12491 & 87855 & 404 & 709 & 0.8\\
\hline
oICF \cite{hilke2016online} & 43.2 & 74.3 & 1.1 & 11.3\% & 48.5\% & 6651 & 96515 & 381 & 1404 & 0.4\\
\hline
MHT\_DAM \cite{kim2015multiple} & 42.9 & \textbf{76.6} & \textbf{1.0} & 13.6\% & 46.9\% & \textbf{5668} & 97919 & 499 & 659 & 0.8\\
\hline
LINF1 \cite{fagot2016improving} & 41.0 & 74.8 & 1.3 & 11.6\% & 51.3\% & 7896 & 99224 & 430 & 963 & 1.1\\
\hline
EAMTT\_pub \cite{sanchez2016online} & 38.8 & 75.1 & 1.4 & 7.9\% & 49.1\% & 8114 & 102452 & 965 & 1657 & \textbf{11.8}\\
\hline
\end{tabular}
\end{center}
\caption{Public tracking results on the MOT16 benchmark test set, $\uparrow$ means higher is better and $\downarrow$ means lower is better. The meaning for evaluation measures is given in \cite{milan2016mot16}. Trackers are sorted by the order of MOTA. We stress the best result under each evaluation measure.}
\label{table:public}
\end{table*}

\begin{table*}
\begin{center}
\begin{tabular}{|c|c|c|c|c|c|c|c|c|c|c|c|}
\hline
Method & MOTA$\uparrow$ & MOTP$\uparrow$ & FAF$\downarrow$ & MT$\uparrow$ & ML$\downarrow$ & FP$\downarrow$ & FN$\downarrow$ & ID Sw.$\downarrow$ & Frag$\downarrow$ & Hz$\uparrow$\\
\hline
\textbf{SOT+MOT} & \textbf{68.6} & 78.8 & 2.1 & \textbf{43.9\%} & 19.9\% & 12690 & \textbf{43873} & 737 & 869 & 0.6\\
\hline
KDNT \cite{yu2016poi} & 68.2 & 79.4 & 1.9 & 41.0\% & \textbf{19.0\%} & 11479 & 45605 & 933 & 1093 & 0.7\\
\hline
POI \cite{yu2016poi} & 66.1 & 79.5 & 0.9 & 34.0\% & 20.8\% & 5061 & 55914 & 805 & 3093 & 9.9\\
\hline
MCMOT\_HDM \cite{lee2016multi} & 62.4 & 78.3 & 1.7 & 31.5\% & 24.2\% & 9855 & 57257 & 1394 & 1318 & \textbf{34.9}\\
\hline
NOMTwSDP16 \cite{choi2015near} & 62.2 & \textbf{79.6} & 0.9 & 32.5\% & 31.1\% & 5119 & 63352 & \textbf{406} & \textbf{642} & 3.1\\
\hline
EAMTT \cite{sanchez2016online} & 52.5 & 78.8 & \textbf{0.7} & 19.0\% & 34.9\% & \textbf{4407} & 81223 & 910 & 1321 & 12.2\\
\hline
\end{tabular}
\end{center}
\caption{Private tracking results on the MOT16 benchmark test set, $\uparrow$ means higher is better and $\downarrow$ means lower is better.}
\label{table:private}
\end{table*}

\begin{table*}
\begin{center}
\begin{tabular}{|c|c|c|c|c|c|c|}
\hline
Method & MOTA$\uparrow$ & MOTP$\uparrow$ & FP$\downarrow$ & FN$\downarrow$ & ID Sw.$\downarrow$ & detector\\
\hline
SOT+MOT(priv) & 67.3 & 81.2 & 5338 & 30266 & 529 & private\\
\hline
MOT(priv) & 67.0 & 81.3 & 5259 & 30626 & 524 & private\\
\hline
SOT+MOT & 41.9 & 77.4 & 4235 & 59757 & 198 & public\\
\hline
SOT(template matching)+MOT &41.7 & 77.8 & 4201 & 59954 & 174 & public\\
\hline
MOT & 38.1 & 78.1 & 5014 & 63121 & 249 & public\\
\hline
\end{tabular}
\end{center}
\caption{Comparison of our tracking results on the MOT16 benchmark training set with different settings, $\uparrow$ means higher is better and $\downarrow$ means lower is better.}
\label{table:train}
\end{table*}

We run an analysis for our appearance model on the MOT16 training dataset, which has no overlap with our appearance model's training data. We randomly sample $3000$ positive and $3000$ negative pairs of ground truth detections from the video MOT16-04. The frame number distance of two detections in a pair is unrestricted, \ie each pair is sampled uniform randomly among the whole video. Our appearance model can give an accuracy of $97.4\%$ for positive pairs, and accuracy $98.5\%$ for negative pairs under the threshold we used in our tracker.

\subsection{MOT16 Challenge evaluation}
Table \ref{table:public} provides a comparison between our algorithm and other state-of-the-art methods using public detections on the MOT16 benchmark test set, and table \ref{table:private} provides a comparison using private detections. Our tracker outperforms other state-of-the-art algorithms using private detections on MOTA, and also has comparable performance with other state-of-the-art algorithms using public detections. Our method achieves the lowest or nearly the lowest FN using both detections, which demonstrate the ability of our SOT algorithm to reduce the number of FN. Our robust appearance model leads to a low number of ID switch.

Table \ref{table:train} provides a comparison of our trackers with different settings on the MOT16 benchmark training set, where the ``training set'' actually plays the role of a validation set. ``SOT+MOT'' is the algorithm we present in this paper, and ``MOT'' is our tracking framework without performing the SOT step. From the table we observe our SOT algorithm can improve our tracker's performance, regardless of the quality of detections. We observe the improvement is more significant based on public detections, because the public detector fail to detect objects more frequently than the private detector, sometimes even fail in simple scenes, resulting in more FNs that our SOT algorithm can recover. We also implemented a naive template matching algorithm for the appearance model used in SOT, which is faster than the neural network based model and easy to implement. from the table we can see even with a simple appearance model, performing the SOT step is still greatly helpful to the whole tracking algorithm.

\begin{figure*}
\begin{center}
\includegraphics[width=\linewidth]{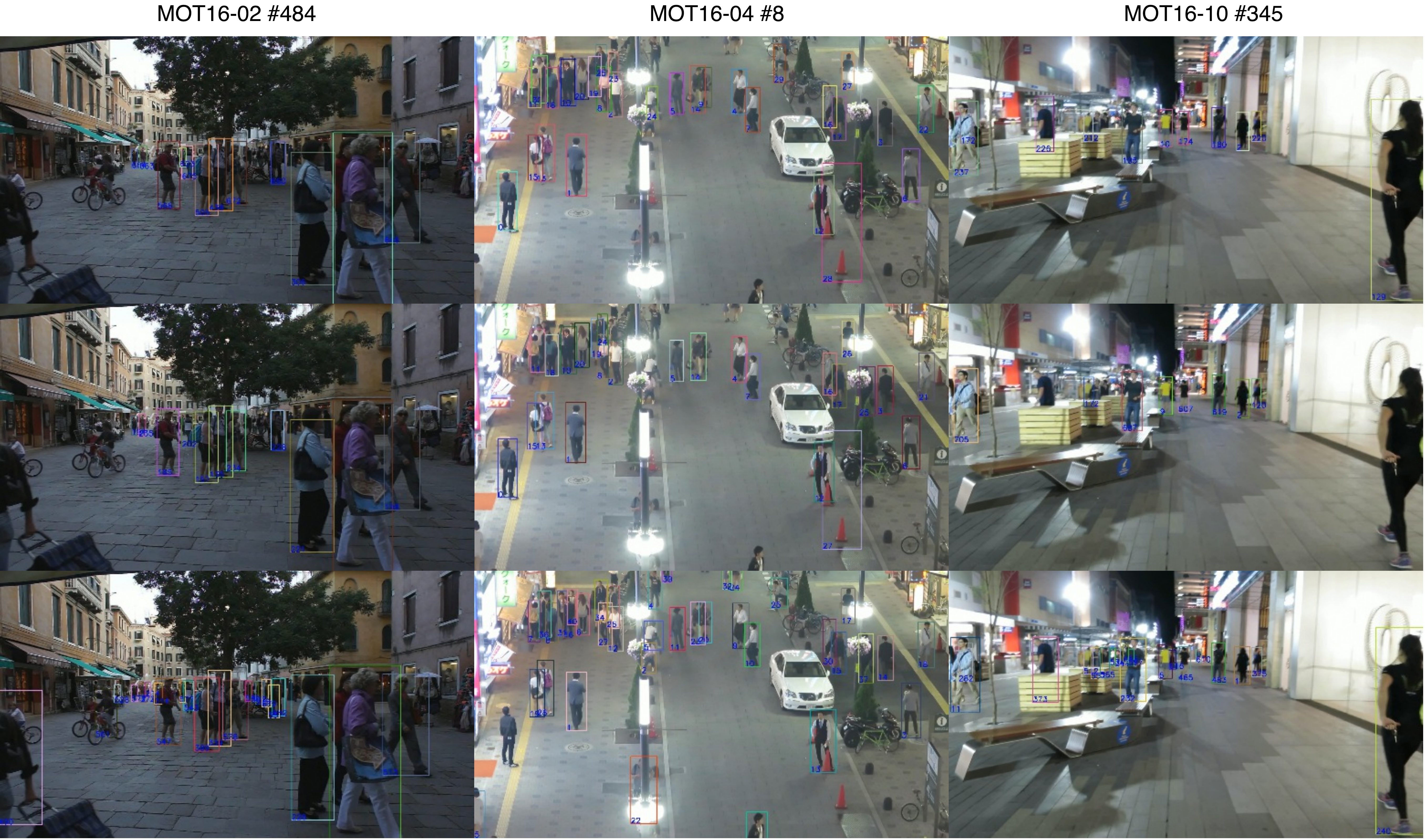}
\end{center}
   \caption{Qualitative examples of our tracking results. The color of bounding boxes and the numbers represent the identity of objects. The first row is our public SOT+MOT results, the second row is our public results without SOT, the third row is our private SOT+MOT results. Our SOT algorithm can help to reduce the number of FNs. The figure is best shown in color.}
\label{Fig:2}
\end{figure*}

Fig \ref{Fig:2} shows some qualitative examples of our tracking results.

\section{Conclusion}
\label{sec:Conclusion}
In this paper, we propose a novel framework that combines SOT algorithm with MOT algorithm, which can significantly reduce the number of FNs while maintaining a low number of FPs. As another contribution, we also propose a deep learning based appearance model that shows strong ability in both detecting missing detections in the SOT part and providing strong affinity measure in the MOT part. In conclusion, experiments show our tracking algorithm runs relatively fast with state-of-the-art performance, which enables the further usage of our algorithm in various applications, like automatic driving and surveillance usages, \etc.

As for future works, we notice that our SOT algorithm can produce useful identity information for detections, \ie it can link detections to tracklets, however we didn't make good use of this information. Another way of possible improvement is the appearance model we used is primitively trained for our MOT algorithm, we directly used it for candidate matching in our SOT algorithm without altering the way of training the model, which may limit the power of our SOT algorithm.

{\small
\bibliographystyle{ieee}
\bibliography{trackingbib}

\begin{thebibliography}{10}\itemsep=-1pt

\bibitem{andriyenko2012discrete}
A.~Andriyenko, K.~Schindler, and S.~Roth.
\newblock Discrete-continuous optimization for multi-target tracking.
\newblock In {\em Computer Vision and Pattern Recognition (CVPR), 2012 IEEE
  Conference on}, pages 1926--1933. IEEE, 2012.

\bibitem{bernardin2008evaluating}
K.~Bernardin and R.~Stiefelhagen.
\newblock Evaluating multiple object tracking performance: the clear mot
  metrics.
\newblock {\em EURASIP Journal on Image and Video Processing}, 2008(1):1--10,
  2008.

\bibitem{bertinetto2016staple}
L.~Bertinetto, J.~Valmadre, S.~Golodetz, O.~Miksik, and P.~Torr.
\newblock Staple: Complementary learners for real-time tracking.
\newblock In {\em International Conference on Computer Vision and Pattern
  Recognition}, 2016.

\bibitem{briechle2001template}
K.~Briechle and U.~D. Hanebeck.
\newblock Template matching using fast normalized cross correlation.
\newblock In {\em Aerospace/Defense Sensing, Simulation, and Controls}, pages
  95--102. International Society for Optics and Photonics, 2001.

\bibitem{butt2013multi}
A.~A. Butt and R.~T. Collins.
\newblock Multi-target tracking by lagrangian relaxation to min-cost network
  flow.
\newblock In {\em Proceedings of the IEEE Conference on Computer Vision and
  Pattern Recognition}, pages 1846--1853, 2013.

\bibitem{choi2015near}
W.~Choi.
\newblock Near-online multi-target tracking with aggregated local flow
  descriptor.
\newblock In {\em Proceedings of the IEEE International Conference on Computer
  Vision}, pages 3029--3037, 2015.

\bibitem{choi2013general}
W.~Choi, C.~Pantofaru, and S.~Savarese.
\newblock A general framework for tracking multiple people from a moving
  camera.
\newblock {\em IEEE transactions on pattern analysis and machine intelligence},
  35(7):1577--1591, 2013.

\bibitem{dalal2005histograms}
N.~Dalal and B.~Triggs.
\newblock Histograms of oriented gradients for human detection.
\newblock In {\em 2005 IEEE Computer Society Conference on Computer Vision and
  Pattern Recognition (CVPR'05)}, volume~1, pages 886--893. IEEE, 2005.

\bibitem{dehghan2015gmmcp}
A.~Dehghan, S.~Modiri~Assari, and M.~Shah.
\newblock Gmmcp tracker: Globally optimal generalized maximum multi clique
  problem for multiple object tracking.
\newblock In {\em Proceedings of the IEEE Conference on Computer Vision and
  Pattern Recognition}, pages 4091--4099, 2015.

\bibitem{dollar2009integral}
P.~Doll{\'a}r, Z.~Tu, P.~Perona, and S.~Belongie.
\newblock Integral channel features.
\newblock In {\em British Machine Vision Conference, BMVC 2009, London, UK,
  September 7-10, 2009. Proceedings}, 2009.

\bibitem{ess2007depth}
A.~Ess, B.~Leibe, and L.~Van~Gool.
\newblock Depth and appearance for mobile scene analysis.
\newblock In {\em 2007 IEEE 11th International Conference on Computer Vision},
  pages 1--8. IEEE, 2007.

\bibitem{fagot2016improving}
L.~Fagot-Bouquet, R.~Audigier, Y.~Dhome, and F.~Lerasle.
\newblock Improving multi-frame data association with sparse representations
  for robust near-online multi-object tracking.
\newblock In {\em European Conference on Computer Vision}, pages 774--790.
  Springer, 2016.

\bibitem{felzenszwalb2010object}
P.~F. Felzenszwalb, R.~B. Girshick, D.~McAllester, and D.~Ramanan.
\newblock Object detection with discriminatively trained part-based models.
\newblock {\em IEEE transactions on pattern analysis and machine intelligence},
  32(9):1627--1645, 2010.

\bibitem{godec2013hough}
M.~Godec, P.~M. Roth, and H.~Bischof.
\newblock Hough-based tracking of non-rigid objects.
\newblock {\em Computer Vision and Image Understanding}, 117(10):1245--1256,
  2013.

\bibitem{he2015deep}
K.~He, X.~Zhang, S.~Ren, and J.~Sun.
\newblock Deep residual learning for image recognition.
\newblock {\em Computer Science}, 2015.

\bibitem{hong2016online}
J.~Hong~Yoon, C.-R. Lee, M.-H. Yang, and K.-J. Yoon.
\newblock Online multi-object tracking via structural constraint event
  aggregation.
\newblock In {\em Proceedings of the IEEE Conference on Computer Vision and
  Pattern Recognition}, pages 1392--1400, 2016.

\bibitem{huang2008robust}
C.~Huang, B.~Wu, and R.~Nevatia.
\newblock Robust object tracking by hierarchical association of detection
  responses.
\newblock In {\em European Conference on Computer Vision}, pages 788--801.
  Springer, 2008.

\bibitem{keuper2016multi}
M.~Keuper, S.~Tang, Y.~Zhongjie, B.~Andres, T.~Brox, and B.~Schiele.
\newblock A multi-cut formulation for joint segmentation and tracking of
  multiple objects.
\newblock {\em arXiv preprint arXiv:1607.06317}, 2016.

\bibitem{khan2005mcmc}
Z.~Khan, T.~Balch, and F.~Dellaert.
\newblock Mcmc-based particle filtering for tracking a variable number of
  interacting targets.
\newblock {\em IEEE transactions on pattern analysis and machine intelligence},
  27(11):1805--1819, 2005.

\bibitem{hilke2016online}
H.~Kieritz, S.~Becker, W.~Hubner, and M.~Arens.
\newblock Online multi-person tracking using integral channel features.
\newblock In {\em 2016 13th IEEE International Conference on Advanced Video and
  Signal Based Surveillance (AVSS)}, pages 122--130, 2016.

\bibitem{kim2015multiple}
C.~Kim, F.~Li, A.~Ciptadi, and J.~M. Rehg.
\newblock Multiple hypothesis tracking revisited.
\newblock In {\em Proceedings of the IEEE International Conference on Computer
  Vision}, pages 4696--4704, 2015.

\bibitem{kuhn1955hungarian}
H.~W. Kuhn.
\newblock The hungarian method for the assignment problem.
\newblock {\em Naval research logistics quarterly}, 2(1-2):83--97, 1955.

\bibitem{kuo2010multi}
C.-H. Kuo, C.~Huang, and R.~Nevatia.
\newblock Multi-target tracking by on-line learned discriminative appearance
  models.
\newblock In {\em Computer Vision and Pattern Recognition (CVPR), 2010 IEEE
  Conference on}, pages 685--692. IEEE, 2010.

\bibitem{leal2015motchallenge}
L.~Leal-Taix\'{e}, A.~Milan, I.~Reid, S.~Roth, and K.~Schindler.
\newblock {MOTC}hallenge 2015: {T}owards a benchmark for multi-target tracking.
\newblock {\em arXiv:1504.01942 [cs]}, Apr. 2015.
\newblock arXiv: 1504.01942.

\bibitem{lee2016multi}
B.~Lee, E.~Erdenee, S.~Jin, and P.~K. Rhee.
\newblock Multi-class multi-object tracking using changing point detection.
\newblock {\em arXiv preprint arXiv:1608.08434}, 2016.

\bibitem{li2014deepreid}
W.~Li, R.~Zhao, T.~Xiao, and X.~Wang.
\newblock Deepreid: Deep filter pairing neural network for person
  re-identification.
\newblock In {\em Proceedings of the IEEE Conference on Computer Vision and
  Pattern Recognition}, pages 152--159, 2014.

\bibitem{liu2016end}
H.~Liu, J.~Feng, M.~Qi, J.~Jiang, and S.~Yan.
\newblock End-to-end comparative attention networks for person
  re-identification.
\newblock {\em arXiv preprint arXiv:1606.04404}, 2016.

\bibitem{liu2016structural}
S.~Liu, T.~Zhang, X.~Cao, and C.~Xu.
\newblock Structural correlation filter for robust visual tracking.
\newblock In {\em Proceedings of the IEEE Conference on Computer Vision and
  Pattern Recognition}, pages 4312--4320, 2016.

\bibitem{milan2016mot16}
A.~Milan, L.~Leal-Taix\'{e}, I.~Reid, S.~Roth, and K.~Schindler.
\newblock {MOT}16: {A} benchmark for multi-object tracking.
\newblock {\em arXiv:1603.00831 [cs]}, Mar. 2016.
\newblock arXiv: 1603.00831.

\bibitem{milan2016online}
A.~Milan, S.~H. Rezatofighi, A.~Dick, K.~Schindler, and I.~Reid.
\newblock Online multi-target tracking using recurrent neural networks.
\newblock {\em arXiv preprint arXiv:1604.03635}, 2016.

\bibitem{milan2014continuous}
A.~Milan, S.~Roth, and K.~Schindler.
\newblock Continuous energy minimization for multitarget tracking.
\newblock {\em IEEE transactions on pattern analysis and machine intelligence},
  36(1):58--72, 2014.

\bibitem{nam2015learning}
H.~Nam and B.~Han.
\newblock Learning multi-domain convolutional neural networks for visual
  tracking.
\newblock {\em Computer Science}, 2015.

\bibitem{pirsiavash2011globally}
H.~Pirsiavash, D.~Ramanan, and C.~C. Fowlkes.
\newblock Globally-optimal greedy algorithms for tracking a variable number of
  objects.
\newblock In {\em Computer Vision and Pattern Recognition (CVPR), 2011 IEEE
  Conference on}, pages 1201--1208. IEEE, 2011.

\bibitem{prokaj2011inferring}
J.~Prokaj, M.~Duchaineau, and G.~Medioni.
\newblock Inferring tracklets for multi-object tracking.
\newblock In {\em CVPR 2011 WORKSHOPS}, pages 37--44. IEEE, 2011.

\bibitem{qi2016hedged}
Y.~Qi, S.~Zhang, L.~Qin, H.~Yao, Q.~Huang, and J.~L. M.-H. Yang.
\newblock Hedged deep tracking.
\newblock In {\em Proceedings of IEEE Conference on Computer Vision and Pattern
  Recognition}, 2016.

\bibitem{ristani2014tracking}
E.~Ristani and C.~Tomasi.
\newblock Tracking multiple people online and in real time.
\newblock In {\em Asian Conference on Computer Vision}, pages 444--459.
  Springer, 2014.

\bibitem{sanchez2016online}
R.~Sanchez-Matilla, F.~Poiesi, and A.~Cavallaro.
\newblock Online multi-target tracking with strong and weak detections.
\newblock In {\em European Conference on Computer Vision}, pages 84--99.
  Springer, 2016.

\bibitem{schroff2015facenet}
F.~Schroff, D.~Kalenichenko, and J.~Philbin.
\newblock Facenet: A unified embedding for face recognition and clustering.
\newblock In {\em Proceedings of the IEEE Conference on Computer Vision and
  Pattern Recognition}, pages 815--823, 2015.

\bibitem{tang2015subgraph}
S.~Tang, B.~Andres, M.~Andriluka, and B.~Schiele.
\newblock Subgraph decomposition for multi-target tracking.
\newblock In {\em Proceedings of the IEEE Conference on Computer Vision and
  Pattern Recognition}, pages 5033--5041, 2015.

\bibitem{tang2016multi}
S.~Tang, B.~Andres, M.~Andriluka, and B.~Schiele.
\newblock Multi-person tracking by multicut and deep matching.
\newblock In {\em European Conference on Computer Vision}, pages 100--111.
  Springer, 2016.

\bibitem{tao2016siamese}
R.~Tao, E.~Gavves, and A.~W. Smeulders.
\newblock Siamese instance search for tracking.
\newblock {\em arXiv preprint arXiv:1605.05863}, 2016.

\bibitem{wang2016tracklet}
B.~Wang, G.~Wang, K.~L. Chan, and L.~Wang.
\newblock Tracklet association by online target-specific metric learning and
  coherent dynamics estimation.
\newblock 2016.

\bibitem{wang2014tracklet}
B.~Wang, G.~Wang, K.~Luk~Chan, and L.~Wang.
\newblock Tracklet association with online target-specific metric learning.
\newblock In {\em Proceedings of the IEEE Conference on Computer Vision and
  Pattern Recognition}, pages 1234--1241, 2014.

\bibitem{wang2015visual}
L.~Wang, W.~Ouyang, X.~Wang, and H.~Lu.
\newblock Visual tracking with fully convolutional networks.
\newblock In {\em Proceedings of the IEEE International Conference on Computer
  Vision}, pages 3119--3127, 2015.

\bibitem{wang2016stct}
L.~Wang, W.~Ouyang, X.~Wang, and H.~Lu.
\newblock Stct: Sequentially training convolutional networks for visual
  tracking.
\newblock CVPR, 2016.

\bibitem{wen2014multiple}
L.~Wen, W.~Li, J.~Yan, Z.~Lei, D.~Yi, and S.~Z. Li.
\newblock Multiple target tracking based on undirected hierarchical relation
  hypergraph.
\newblock In {\em 2014 IEEE Conference on Computer Vision and Pattern
  Recognition}, pages 1282--1289. IEEE, 2014.

\bibitem{yang2011learning}
B.~Yang, C.~Huang, and R.~Nevatia.
\newblock Learning affinities and dependencies for multi-target tracking using
  a crf model.
\newblock In {\em Computer Vision and Pattern Recognition (CVPR), 2011 IEEE
  Conference on}, pages 1233--1240. IEEE, 2011.

\bibitem{yang2012online}
B.~Yang and R.~Nevatia.
\newblock An online learned crf model for multi-target tracking.
\newblock In {\em Computer Vision and Pattern Recognition (CVPR), 2012 IEEE
  Conference on}, pages 2034--2041. IEEE, 2012.

\bibitem{yu2016poi}
F.~Yu, W.~Li, Q.~Li, Y.~Liu, X.~Shi, and J.~Yan.
\newblock Poi: Multiple object tracking with high performance detection and
  appearance feature.
\newblock {\em arXiv preprint arXiv:1610.06136}, 2016.

\bibitem{zamir2012gmcp}
A.~R. Zamir, A.~Dehghan, and M.~Shah.
\newblock Gmcp-tracker: Global multi-object tracking using generalized minimum
  clique graphs.
\newblock In {\em Computer Vision--ECCV 2012}, pages 343--356. Springer, 2012.

\bibitem{zhang2008global}
L.~Zhang, Y.~Li, and R.~Nevatia.
\newblock Global data association for multi-object tracking using network
  flows.
\newblock In {\em Computer Vision and Pattern Recognition, 2008. CVPR 2008.
  IEEE Conference on}, pages 1--8. IEEE, 2008.

\bibitem{zheng2015scalable}
L.~Zheng, L.~Shen, L.~Tian, S.~Wang, J.~Wang, and Q.~Tian.
\newblock Scalable person re-identification: A benchmark.
\newblock In {\em Proceedings of the IEEE International Conference on Computer
  Vision}, pages 1116--1124, 2015.

\bibitem{zhu2016beyond}
G.~Zhu, F.~Porikli, and H.~Li.
\newblock Beyond local search: Tracking objects everywhere with
  instance-specific proposals.
\newblock {\em arXiv preprint arXiv:1605.01839}, 2016.

\end{thebibliography}
}

\end{document}